\documentclass[10pt,leqno]{amsart}
\usepackage{graphicx}
\usepackage{indentfirst,csquotes}

\usepackage[T1]{fontenc}
\usepackage{multirow}
\usepackage{graphicx}
\usepackage{amssymb,amsthm,amsmath}
\usepackage{xcolor}
\usepackage{hyperref}
\hypersetup{ colorlinks=true, linkcolor=black, filecolor=black, urlcolor=black }

\usepackage{lipsum}

\begin{document}
\title{Predicting clinical outcomes\\ from patient care pathways\\ represented with temporal knowledge graphs}
%
%
 \author{Jong Ho Jhee$^{1}$ 
 \and
 Alberto Megina$^{1}$ \and 
 Pacôme Constant Dit Beaufils$^{2,3}$
 \and
 Matilde Karakachoff$^{3,4}$
 \and 
 Richard Redon$^{2}$
 \and 
 Alban Gaignard$^{2}$
 \and 
 Adrien Coulet$^{1}$
 }

%
 \address{Affiliations: 1. Inria, Inserm, Université Paris Cité, HeKA, UMR 1346, Paris, France\\ --
 2. CNRS, Inserm, Université de Nantes, Institut du Thorax, UMR 1087, Nantes, France\\ --
 3. Université de Nantes, CHU Nantes, Nantes, France\\ --
 4. Inserm, Clinique des données, CIC 1413, Nantes, France\\
}

 \email{
 1. \{jong-ho.jhee, alberto.megina, adrien.coulet\}@inria.fr\\ -- 
 2. \{richard.redon, alban.gaignard\}@univ-nantes.fr \\ -- 
 3. \{pacome.constantditbeaufils, matilde.karakachoff\}@chu-nantes.fr
 }
\maketitle              
\begin{abstract}
\ \\ \textbf{Background:} 
With the increasing availability of healthcare data, predictive modeling finds many applications in the biomedical domain, such as the evaluation of the level of risk for various conditions, which in turn can guide clinical decision making. 
However, it is unclear how knowledge graph data representations and their embedding, which are competitive in some settings, could be of interest in biomedical predictive modeling. 
\textbf{Method:} 
We simulated synthetic but realistic data of patients with intracranial aneurysm and experimented on the task of predicting their clinical outcome. We compared the performance of various classification approaches on tabular data versus a graph-based representation of the same data. Next, we investigated how the adopted schema for representing first individual data and second temporal data impacts predictive performances. 
\textbf{Results:}
Our study illustrates that in our case, a graph representation and Graph Convolutional Network (GCN) embeddings reach the best performance for a predictive task from observational data. We emphasize the importance of the adopted schema and of the consideration of literal values in the representation of individual data. Our study also moderates the relative impact of various time encoding on GCN performance. 

\end{abstract}
\section{Introduction}
\label{sec:intro}
Intracranial aneurysms are abnormal bulges in the blood vessels of the brain that pose significant health risks particularly when they rupture, leading to severe neurological damage or death \cite{keedy2006overview}. The ability to predict the evolution, and the best management of intracranial aneurysms, especially in the context of their various treatments, is crucial for improving patient outcomes. This predictive capability is particularly important in a clinical setting where understanding the relationship between patient features, the treatments they receive, and their subsequent outcomes can serve as a basis for early or preventive interventions.
In this context, we aim at establishing a method for identifying patients who are at a higher risk of adverse outcomes following intracranial aneurysm rupture. 
By combining patients' personal features and the treatments observed during their hospital stays, we hope to uncover patterns that lead to more accurate predictions of patient outcomes. This would not only enhance the understanding of the effectiveness of different disease management, but also provides a foundation for the development of targeted intervention strategies that could mitigate adverse outcomes.

Graph embeddings are mapping of the nodes and edges that compose a graph into a continuous vector space \cite{hamilton2017representation}. This transformation allows for such complex relational data to be represented in a form that is more amenable to computational analysis. In particular, it allows to apply machine learning algorithms to perform tasks such as link prediction, node classification or clustering with high performance \cite{wang2017knowledge}.
In this work, we especially consider embeddings on Knowledge Graphs (KGs), as defined in the context of the Semantic Web \cite{HBC21}. 
The atomic elements composing those are triples of the form of $\langle \textit{subject}, \textit{predicate}, \textit{object} \rangle$, where the subject and object are nodes of the graph, representing entities; and the predicate is a labelled and oriented edge, representing that a particular relationship stands between the subject and object \cite{FOST}. KG can be encoded in Resource Description Framework (RDF) a standard where entities and predicates are uniquely identified with a Uniform Resource Identifier (URI), facilitating interoperability across different datasets and ontologies. 

However, it is unclear how KG representations and KG embeddings may be of interest in node classification task to predict clinical outcomes \cite{cui2023a}, and in particular to predict outcomes of intracranial aneurysm management. 
We propose in this work to advance some initial answers, by exploring three key scientific questions that guided our investigation.
First, we wonder if the use of various graph embedding approaches is competitive with regard to classical predictive approaches on tabular data. Second, we wonder how these approaches are impacted by modeling choices made for representing individual data in the form of a graph. Third, we wonder about the impact of various possible representation of time, \textit{i.e.} timestamps, sequential relations or both on predictive performances.

The contributions of this article are: 
(\textit{i}) A publicly shared synthetic but realistic dataset of pathways of patients treated for a ruptured intracranial aneurysm; along with scripts that transform data from a tabular form to various graph representations;
(\textit{ii}) Elements of answer to our three questions, illustrating that, in our context, graph embeddings learned with Graph Convolutional Network (GCN) outperform other approaches, that the more compact representation of patient features is associated with better performance and that the representation of time does not impact prediction performances.

The remainder of this article introduces related works in section \ref{sec:related work}, material and methods in section \ref{sec:mat_and_met}, empirical results and their interpretation in section \ref{sec:experiments} and finally presents a discussion in section \ref{sec:discussion}.


\section{Related work}
\label{sec:related work}

\subsection{Standard schema for clinical data} 
Several data schema have been proposed to model, harmonize and facilitate the exchange of clinical data. For example, FHIR~\cite{lehne2019use} and the OMOP CDM~\cite{voss2015feasibility} are standards proposed to tackle clinical data interoperability. FHIR is loosely specified, which makes it hard to associate this schema with precisely defined semantics. Even if RDF transformations of FHIR \cite{PRUDHOMMEAUX2021103755} and OMOP CDM have been proposed, none has been widely adopted yet~\cite{chytas2024mapping,XIAO2022104201}. 
Similarly, Phenopackets~\cite{jacobsen2022ga4gh} is a uniform data structure to ease the combination and exchange of genomic data and clinical observations. If some works have been proposed to align Phenopackets with semantic web technologies~\cite{Kaliyaperumal2022PhenopacketsFT}, it is insufficient to represent the full spectrum of clinical data. 

\begin{figure}[ht!]
    \centering
    \includegraphics[width=0.95\textwidth]{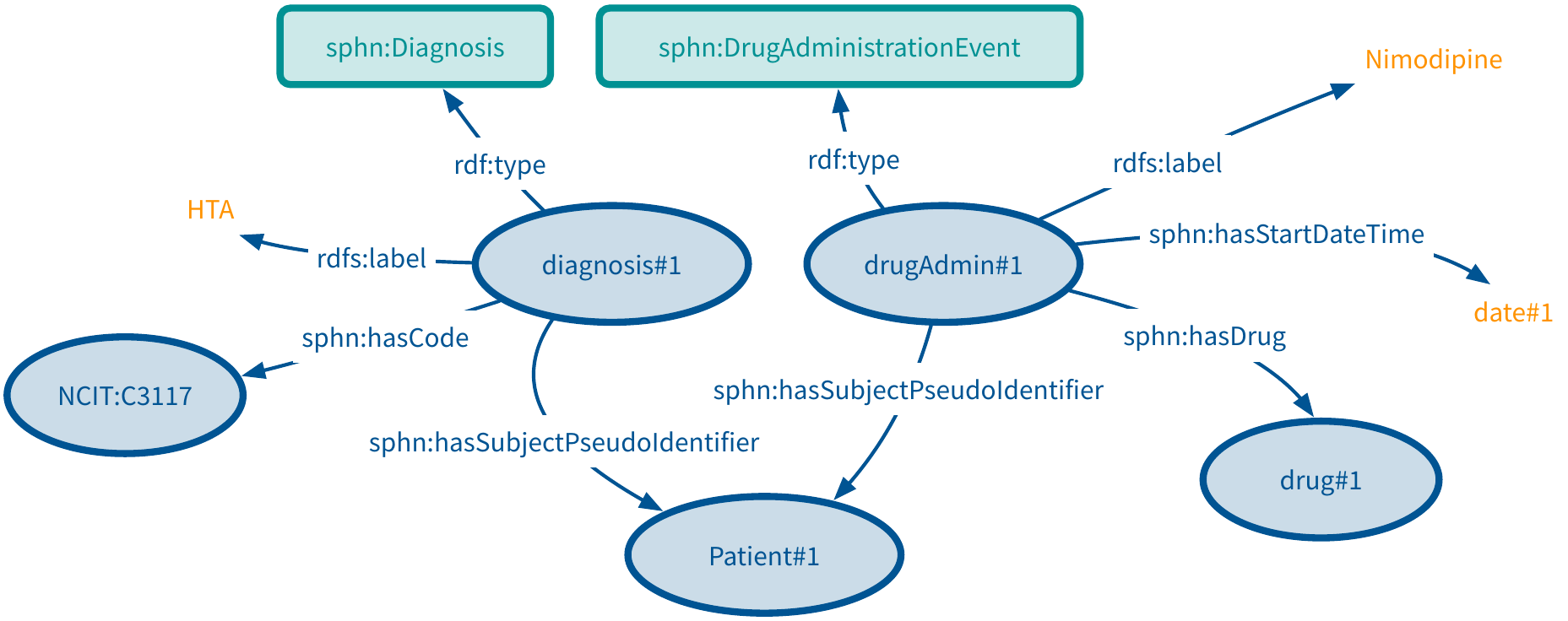}
    \caption{Example of individual clinical data represented in a SPHN knowledge graph. Temporal information is specified with RDF literals associated to events through \texttt{sphn:hasStartDateTime} properties.
    }
    \label{fig:sphn}
\end{figure}

Beside, two ontologies have recently been proposed to represent individual clinical data in the form of knowledge graphs. The first is the SPHN (Swiss Personalized Health Network) ontology~\cite{Tour2023FAIRificationOH}, 
which has been adopted by the five Swiss academic hospitals for better data sharing and integration. As illustrated in Figure~\ref{fig:sphn}, in this semantic model, patient is a central entity, that can be associated with diagnoses, drug administrations and procedures, each of which can potentially be associated with starting and ending times with literals. 

\begin{figure}[ht!]
    \centering
    \includegraphics[width=0.95\textwidth]{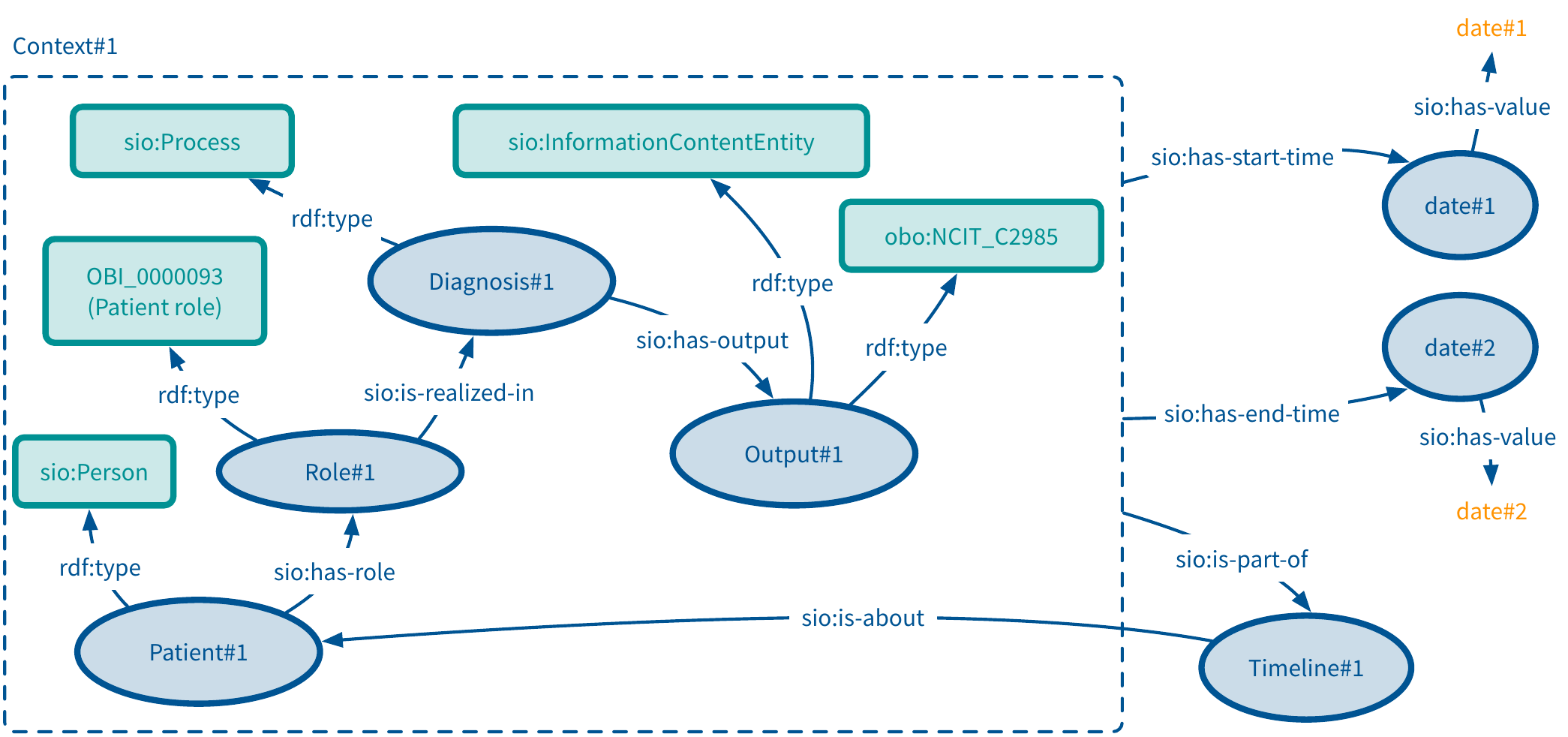}
    \caption{Example of individual clinical data represented in a CARE-SM KG. Temporal information is directly linked to the Context\#1 named graph with start and end dates. Multiple events can be associated to a given timeline through the \texttt{sio:is-part-of} property. 
    }
    \label{fig:care-sm}
\end{figure}

The second is the CARE-SM (Care and Registry Semantic Model) ontology~\cite{Kaliyaperumal2022}. It was initially designed to represent clinical data in the context of rare diseases and largely relies on the reuse of the Semantic Science Integrated Ontology (SIO)~\cite{Dumontier2014TheSI}. As depicted in Figure~\ref{fig:care-sm}, it provides fine-grained representations for the multiple roles with which a person can participate in the context of a care procedure or a research study. The originality of CARE-SM resides in its use of RDF quads. They are used to associate a semantic context to each data element through RDF named graph. This particularly enables the representation of provenance metadata for clinical observations or diagnosis. Notably, these named graphs can be used to represent timelines of clinical events.  

In this work, we focus on SPHN and CARE-SM first because both of them provide sufficiently precise specification to let one represent an arbitrary clinical dataset in the terms of their ontologies, which is not the case of FHIR-RDF; second because they propose very different modeling choices to represent individual clinical data; and third because they are adopted in large scale projects.

\subsection{Time in knowledge graphs} 

The OWL-Time is a standard ontology that provides classes, predicates and patterns to represent time, duration and intervals~\cite{OWLTime2022}. In particular, it enables one to instantiate relations between events with the Allen's interval algebra and can be associated with time reasoning mechanisms \cite{batsakis2017temporal,zhang2021rdf}. Both SPHN and CARE-SM enable various way to represent time, including the use of OWL-Time and Allen's interval algebra. In practice, one may want to restrict themselves to absolute time by only associating timestamps to events, whereas another might want relative time by instantiating relationships such as ``is before'' between events, or even to use both absolute and relative relationships. In addition, if one choose relative time, they also have to decide on a \textit{level of saturation} between events going from a simple sequence, where only directly subsequent events are related, to a fully saturated level where each event is timely related to every other event. 
Even if it is clear that a fully saturated graph is associated with several drawbacks (\textit{e.g.}, the high connectivity of temporal nodes makes the exploration of the graph complex), the most adapted modeling of time to adopt for a particular task is not always clear~\cite{zhang2021rdf}.

\subsection{KG embeddings and node classification}

Many approaches for representing entities of a KG within a latent space exist, and categorisations of them have been proposed in \cite{cai2018comprehensive}. 
As an illustration, TransE \cite{BUGD13} is an embedding approach that works at the triple level as it aims at minimizing the following scoring function: 
\[
f_r(h, t) = \| \mathbf{h} + \mathbf{r} - \mathbf{t} \|_{1} ,
\] 
where $\mathbf{h}$, $\mathbf{r}$, and $\mathbf{t}$ are the embedding vectors of the head entity, relation, and tail entity, respectively. Here the relationship between the head and the tail (\textit{i.e.}, the subject and object) can be seen as a translation $r$ in the embedding space. 
RDF2Vec follows a very different approach that works at the sequence level~\cite{ristoski2016rdf2vec}. It uses random walks drawn from the knowledge graph. These sequences of either edges, nodes, or subtrees are used to feed a word2vec model that outputs embeddings for each node in a sequence, for example, by maximizing the probability of a node given the previous nodes of the sequence. For instance, the Continuous Bag-of-Words model (CBOW), one of the algorithms associated with word2vec, maximizes the average log probability of the target node given a sequence of nodes:
\[
\frac{1}{T} \sum_{t=1}^{T} \log{p(e_t|e_{t-c},...,e_{t+c}}) ,
\]
where $e_t$ is a target node and $c$ denotes the context window. In contrast with TransE and RDF2Vec, which work at the triple and sequence levels, GCNs \cite{kipf2017} work at the level of the neighborhood of nodes. They have been introduced for classification over graphs and extended for node classification and link prediction in knowledge graphs \cite{schlichtkrull2018modeling}. GCNs compute the embeddings of a node by considering its neighborhood in the graph. GCNs can be seen as a message-passing framework of multiple layers, in which the embedding $h^{(l+1)}_i$ of a node $i$ at layer $(l+1)$ depends on the embeddings of its neighbors at layer $(l)$, as follows:
\[
h_i^{(l+1)} = \sigma \Bigg( \underbrace{\sum_{j \in \mathcal{N}_i} \frac{1}{c_{i}} W^{(l)} h_j^{(l)}}_{\text{Neighborhood}} +  \underbrace{W^{(l)} h_i^{(l)}}_{\text{Self-connection}} \Bigg) .
\]
The convolution over the neighboring nodes $j$ of $i$ is computed with a learnable weight matrix $W^{(l)}$ and each layer $(l)$; and is normalized by a constant $c_{i}$. The last term enables self-connection \textit{i.e.}, the fact that the embeddings of the node $i$ at a particular layer $(l+1)$ also depends on its embeddings at the layer $(l)$. Relational Graph Convolutional Networks (RGCN) \cite{schlichtkrull2018modeling} is a particular type of GCNs that takes into consideration multi-relational data, \textit{i.e.}, considers differently a same neighbor node when related through different labelled edges to $i$. This is particularly adapted to the semantic web KGs since they encompass various predicates to represent relations associated with different semantics. To this aim, RGCNs incorporate entity embeddings with multi-relations in the neighborhood aggregation scheme, as follows:
\[
h_i^{(l+1)} = \sigma \Bigg(\sum_{r \in \mathcal{R}} \sum_{j \in \mathcal{N}_i^r} \frac{1}{c_{i,r}} W_r^{(l)} h_j^{(l)} + W_0^{(l)} h_i^{(l)} \Bigg) .    
\]
where the convolution over the neighbor nodes is computed using a specific weight matrix $W^{(l)}_r$ for each predicate (relation type) $r \in \mathcal{R}$; and $c_{i,r}$ is a normalizing constant such as the number of neighboring nodes $|\mathcal{N}_i^r|$. The parameter sharing and sparsity constraints using basis-decomposition allow RGCN to deal with the large number of relations. In the following, we consider TransE, RDF2Vec as two baseline approaches for KG embedding (KGE) models and RGCN as it seems a well adapted candidate to take into consideration the variety of relationship types of our clinical KG. 

Values associated with entities through the use of literals are generally disregarded by KGE approaches. However various works investigates the interest of incorporating them to node representations. To this aim LiteralE~\cite{kristiadi2019incorporating}, proposes an approach where two vectors represent a single node, the first representing the node embedding itself, and the second containing each of the numerical literal it is associated with. Both vectors are then combined before to be passed to a scoring function. Knowledge Embedding with Numbers (KEN) \cite{cvetkov2023relational} uses an encoder, a single neural network, to inject the literal values into the same vector space with entities. Other approaches have been proposed to consider textual literals or combinations of various type of literals \cite{gesese2021survey}.

One of the main interest of graph embedding is to facilitate the application of machine learning algorithms on complex relational data to perform a variety of tasks such as link prediction, node classification, graph classification or node clustering \cite{HBC21}.
In this work we consider three embedding approaches, but only one subsequent task that is node classification. 
This learning task can be defined as the estimation of the likelihood that a given entity, not explicitly asserted in a KG, is assigned to a specific type \cite{wang2017knowledge}. Using KG embeddings, this task relies on the assumption that the embedding space effectively captures the inherent characteristics and structural properties of the original graph. In practice, the spatial configuration of vectors should reflect the relational information observed in the KG, thus allowing predictions about node labels. This task is formally described in the Materials and Methods section of this article.

\subsection{Temporal knowledge graph embeddings}
A temporal knowledge graph (TKG) extends existing KG by incorporating temporal information in a specific time or interval. This addition poses significant challenges because it requires integrating the temporal validity of facts into the models in order to accurately capture the dynamics of entities and relations over time \cite{dall2024embedding}. Effectively modeling the temporal aspect is crucial for applications where the evolution of relations is significant, such as in clinical data where the time of treatments, procedures and the patient state can critically influence outcomes. Temporal knowledge graph embedding models use triples tagged with each timestamp, which are quadruples, and this is considered as learning the representation of each time-snapshot graph. However, the time interval could be critical information, or the graph might not be compatible with quadruple forms, as not all triples necessarily possess temporal information. Encoding or transforming temporal information in the form of vectors of literals or relations can be applied in this type of graph \cite{cai2023temporal}. 

\section{Materials and Methods}
\label{sec:mat_and_met}

\subsection{Synthetic data generation}
We built a synthetic data set of 10,000 patients with 30 clinical features and 1 outcome variable. Out of the 30 clinical features, 8 are associated with a timestamp and for this reason are hereafter named events, to distinguish from demographic or historical features not associated with a particular time. In our dataset, events have the particularity of being associated with only one timestamp, which is the time between hospital admission and the first occurrence of the event during the patient's stay. Table~\ref{tab:list_features} in Appendix lists the 22 clinical features (non-temporal) and 8 events of our synthetic dataset. 


 
To make our synthetic dataset as realistic as possible, patient features and care events were generated according to observations made on a real-world clinical dataset of 552 patients diagnosed with a ruptured intracranial aneurysm, provided by the Nantes 
University Hospital. Access to this dataset has been granted under an IRB agreement. 

\begin{figure}[t]
    \centering
    \includegraphics[width=1.0\textwidth]{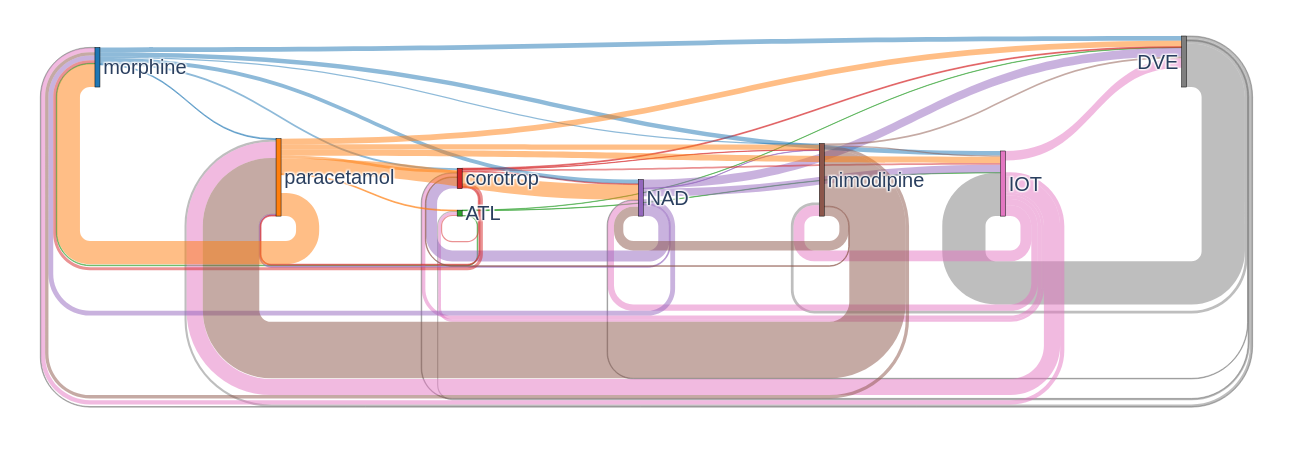}
    \caption{A visual representation of care pathways where the larger connections between care events correspond to the higher transition probabilities. {\tt morphine}: morphine use, {\tt paracetamol}: paracetamol use, {\tt corotrop}: milrinone use, {\tt ATL}: percutaneous transluminal angioplasty, {\tt NAD}: norepinephrin use, {\tt nimodipine}: nimodipine use, {\tt IOT}: orotracheal intubation, and {\tt DVE}: external ventricular drainage.}
    \label{fig:care-events}
\end{figure}

First we estimated the closest distribution for each feature of the real-world dataset using a Kolmogorov–Smirnov test. For instance, we observed that the duration of patients' hospital stay was following a generalized extreme value distribution. Second, after performing a factorization for categorical variables, we computed correlations across each possible pair of features, and computed the transition probabilities across types of events. We observed that the duration of hospital stay was highly correlated with the number of medical procedures received; and a high probability for Paracetamol to be administered subsequently to Nimodipine. Figure \ref{fig:care-events} shows a Sankey diagram built from the observed transition probabilities, which visually summarizes the possible care pathways. To simplify the representation of time in our dataset, the 8 events were binarized by a transformation into pairs of events set to one if the patient observed a transition between the first and second event, and to zero otherwise.
After validation by clinical experts of the consistency of the distributions, correlations and transition probabilities, we used them to constrain the generation of our synthetic dataset.

Finally, we generated the patient outcome feature with one of the three distinct values \{\textit{BackHome, Rehabilitation, Death}\} associated with the following proportions in the synthetic dataset: 44.14\%, 43.33\% and 12.53\%, respectively, mirroring the real-word distribution. 


\subsection{Graph representation of clinical data}
\label{subsec:graph-rep-of-clinic}
We developed transformations of our synthetic tabular dataset using various modeling choices. 
By using RDF templates and rules, we generated a set of graphs that instantiate the SPHN ontology: 
\begin{itemize}
    \item 
{\tt SPHN-nl} (no literals) where all the literals were removed; 
    \item 
{\tt SPHN-nt} (no time) where temporal information were removed; 
    \item 
{\tt SPHN-ts} (timestamps) with timestamps only;
    \item 
{\tt SPHN-tr} (time relations) with a single \texttt{time:before} predicate between directly subsequent events;
    \item 
{\tt SPHN-tsr} (timestamps and relations) with both timestamps and relations between subsequent events;
    \item 
{\tt SPHN-sat1} (saturation level 1) where {\tt SPHN-tr} is enriched with additional \texttt{time:before} predicates obtained by applying once the following transitivity rule  $\texttt{time:before} \circ \texttt{time:before} \sqsubseteq \texttt{time:before}$;
    \item 
{\tt SPHN-sat2} (saturation level 2) by applying the same transitivity twice.
\end{itemize}

\begin{figure}[t]
    \centering
    \includegraphics[width=0.80\linewidth]{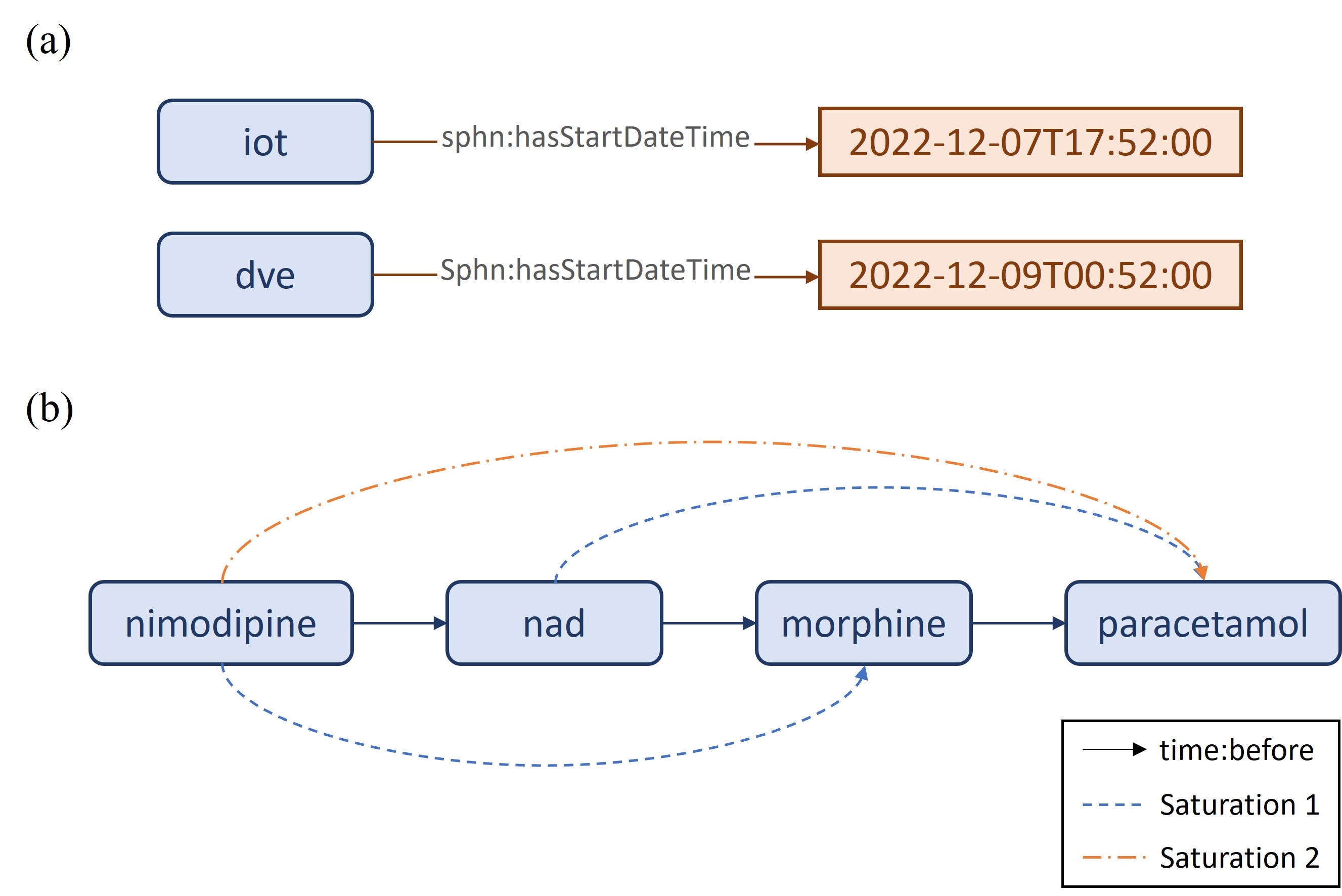}
    \caption{Examples of temporal information: (a) two event associated with a timestamp; (b) sequence of events related with \texttt{time:before} relations. In plain lines are relations between directly subsequent events. Applying a transitivity rule once, add 2 relations in dashed doted line (saturation 1) and twice, add the last relation depicted with the dash-dotted line (saturation 2). {\tt iot}: orotracheal intubation, {\tt dve}: external ventricular drainage, {\tt nad}: nicotinamide adenine dinucleotide.}
    \label{fig:temporal-info}
\end{figure}

\noindent Figure \ref{fig:temporal-info} illustrates timestamps and time relation between events, at three level of saturation corresponding to {\tt SPHN-tr}, {\tt SPHN-sat1} and {\tt SPHN-sat2}.

By using RDF templates, we also generated a graph that instantiate CARE-SM ontology. In this case, we performed an additional step where quads were transformed into triples, in particular by using a \texttt{nvasc:hasTimePoint} link to associate directly instances such as diagnoses or drug administrations to their absolute time. 
This step was motivated by the fact that KGE approaches do not support quads, but only triples. The resulting graph is named \textit{CARE-SM$^*$}. We generated the same variants of graphs as we did for SPHN, but report only here about {\tt CARE-SM$^*$-ts} that is the version with timestamps only. Table \ref{tab:dataset} in Appendix shows the statistics of SPHN and CARE-SM$^*$ graphs.



We make two last transformations to each variant of our graphs: first, to ensure that the orientation of predicates do not influence our experiments, inverse relationships were systematically asserted; 
we encoded timestamp literals into continuous numbers in the range of [0, 1] using a quantile transformation. This spreads out the most frequent values thus reduces the impact of outliers \cite{ehm2016quantiles}.
Scripts for these various transformation are provided at {\small \href{https://github.com/TeamHeka/neurovasc}{\small \texttt{https://github.com/TeamHeka/neurovasc}}}.

\subsection{KG embedding and patient outcome prediction}

The objective of our prediction task is to forecast patient outcomes based on their clinical features and experienced events. 
In the case of the graph dataset, we model this task as a node classification problem aiming at associating patient nodes with the class that corresponds to their correct outcome.
\subsubsection{RGCN for patient outcome prediction} 
\begin{figure}[t]
    \centering
    \includegraphics[width=0.95\textwidth]{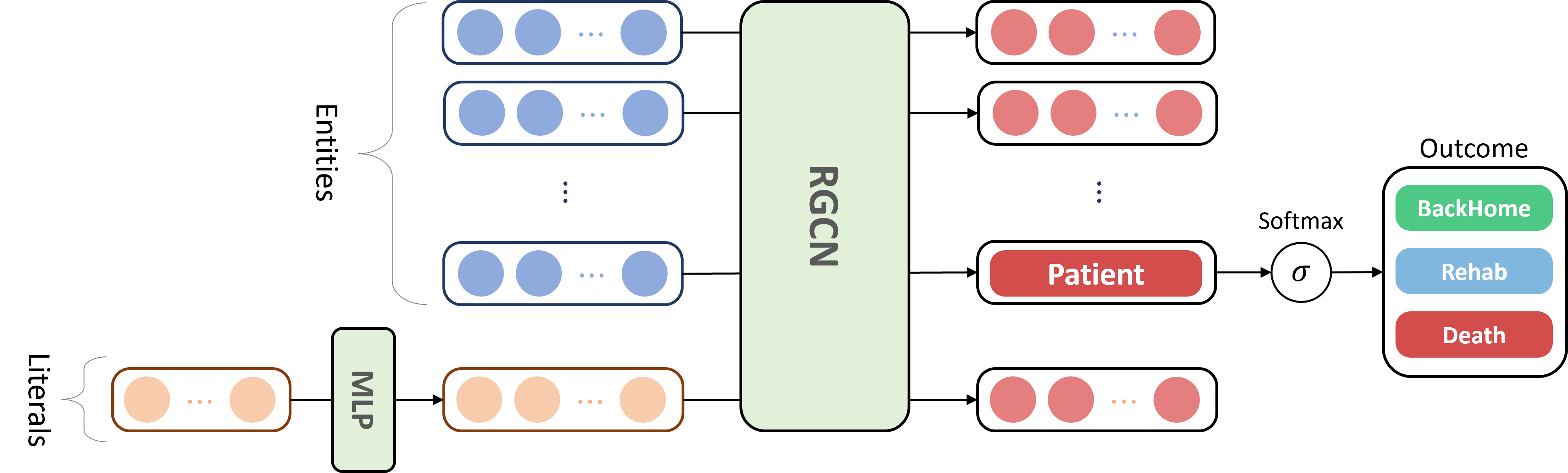}
    \caption{An illustration of the model denoted RGCN+lit for patient outcome prediction.}
    \label{fig:rgcn}
\end{figure}

In this section, we describe in detail the graph embedding model, RGCN$+$Literals for the outcome prediction. The overall architecture is illustrated in Figure \ref{fig:rgcn}. Given a graph $G = ( \mathcal{V}, \mathcal{E}, \mathcal{R}, \mathcal{X} )$, where $\mathcal{V}$ denotes the set of nodes (entities), $\mathcal{E}$ the set of edges, $\mathcal{R}$ the set of relations (predicates), and $\mathcal{X} \in \mathbb{R}^{n \times d_0}$ denotes the initial input embeddings of dimension $d_0$. The representation of a target node (patient) $h_i^{(1)} \in \mathbb{R}^{d_1}$ for $i \in |\mathcal{V}|$ after passing a first RGCN layer is defined as:
\[
h_i^{(1)} = \sigma \Bigg(\sum_{r \in \mathcal{R}} \sum_{j \in \mathcal{N}_i^r} \frac{1}{c_{i,r}} W_r^{(0)} x_j + W_0^{(0)} x_i \Bigg) ,
\]
where $x_i \in \mathcal{X}$ is an initial patient embedding, $W_r \in \mathbb{R}^{d_o \times d_1}$ denotes a weight matrix for each relation $r$ and $\sigma$ is a non-linear activation function. Additionally, the number of parameters increases with the number of relations, which can lead to overfitting on some of the rare relations. Thus we apply basis-decomposition method for the regularization of the model \cite{schlichtkrull2018modeling}.

The final patient node embedding is extracted after passing $L$ stacks of RGCN-layers and the softmax function is applied to output the probability of outcomes. The model is trained by minimizing the cross-entropy loss on the patient nodes:  
\[
\mathcal{L} = - \sum_{p \in \mathcal{P}} \sum_{k=1}^{K} y_{pk} \log{z_{pk}} ,
\]
where $\mathcal{P}$ is the set of patient nodes in the training set, $K$ the number of outcomes and $z_{pk}$ the probability of the outcome.

\subsubsection{RGCN with Literals} In our clinical KG, some clinical features are in the form of literals (\textit{e.g.}, the age of a patient). However, RGCN model only consider entity nodes and relations, and thus do not take literals into account. 
To mitigate this problem, we propose a model denoted RGCN$+$Literals (or RGCN+lit for short) that employs an additional function for the literals. Before the input of initial embeddings to RGCN, the function of a multi-layer perceptron (MLP) is used to transform the value of the literal into embeddings:
\[
x_{literal} = \sigma (Wv + b) ,
\]
where $v$ denotes the attribute value, $\sigma$ the non-linear activation function, $W$ the weight matrix and $b$ the bias. 
Note that other encoding functions could be applied instead of the MLP. 

\section{Experimental setting and results}
\label{sec:experiments}
In the experiments, we compare various outcome prediction approaches along three distinct experiments on our synthetic dataset. Each experiment is repeated ten times for each model to assess the variability of the 
performance. 
Each time the 10,000 patients are randomly split into train (80\%), validation (10\%) and test (10\%) set. 
Datasets and codes are available at \href{https://github.com/TeamHeka/neurovasc}{\small \url{https://github.com/TeamHeka/neurovasc}}. The three experiments compare:

\begin{itemize}
    \item \textbf{Tabular \textit{vs}. Graph data.} 
    We compare the performance of standard predictive approaches applied on tabular data with those of various KGE approaches to examine whether the multi-relational information within the graph would benefit the predictive task.
    \item \textbf{SPHN \textit{vs.} CARE-SM ontology.} 
    We compare the impact on the prediction of using either one schema or another to evaluate whether the structure of the graph affects the performance. 
    \item \textbf{Various time modeling.} 
    We compare the impact on the prediction of using various time modeling, such as no time data,  absolute time only, relative time only, both absolute and relative time and different levels of saturation of the relative time relationships that connect temporal events.
\end{itemize}

\subsection{Tabular \textit{vs.} Graph data}
We considered the following methods to establish baseline performance from tabular data, Logistic Regression (LR), Random Forest (RF), Feed-forward Neural Network (NN) to compare to KGE approaches. 
RF were set up with 100 trees, 
and NN with three layers with hidden dimension sizes of [100, 50, 10] and hyperbolic tangent was used as an activation function. 

For this first comparison with various KGE approaches, we arbitrarily chose to only consider the SPHN ontology. In addition, to enable a fair comparison with tabular data (which encode only the sequence of event), {\tt SPHN-tr} is considered as the comparative SPHN graph.
To represent three main families of KGE we considered TransE, RDF2Vec and RGCN+lit. All models used an initial embedding dimension of 100. For TransE, 1-norm is applied for the regularization of the scoring function. For RDF2Vec, ten walks of a maximum depth of three for each node was applied using the random walk strategy. The patient representations obtained from the first two KGE approaches are input to a NN model as a classifier. For RGCN+lit, three RGCN layers are applied to aggregate the information within the three-hop neighborhood of patient nodes and Parametric Rectified Linear Unit (PReLU) \cite{he2015delving} is used as a non-linear activation function. And all models were optimized using Adam optimizer \cite{kingma2014adam} with the learning rate of 1e-3 and the weight decay of 5e-4.

The obtained performance is shown in Table \ref{tab:tabgraph-comparison}. All three baseline approaches from tabular data reach poor performances (F1-score = [0.44,0.49], AUC=[0.63, 0.71]). RF showed the best performance (AUC=0.71), closely followed by LR. Especially, the two models showed decent F1-score for {\em BackHome} outcome, compare to other models. For graph data, neither TransE nor RDF2Vec seem to succeed in predicting outcomes (AUC=[0.49,0.5]).  However, the RGCN+lit model showed the overall best result  (F1=0.78, AUC=0.91). 




\begin{table}[ht]
\caption{The patient outcome prediction comparison on Tabular and Graph ({\tt SPHN-tr}). RGCN3+lit refers to RGCN+lit with 3 layers. }
\label{tab:tabgraph-comparison}
\resizebox{\textwidth}{!}{%
\def\arraystretch{1.5}
\begin{tabular}{llccccccc}
\hline
\multirow{2}{*}{\textbf{Type}} & \multirow{2}{*}{\textbf{Model}} & \multicolumn{5}{c}{\textbf{F1-score}}                          & \multirow{2}{*}{\textbf{Accuracy}} & \multirow{2}{*}{\textbf{AUC}} \\ \cline{3-7}
                               &                                 & BackHome  & Rehab          & Death     & Macro     & Weighted  &                                    &                               \\ \hline
\multirow{3}{*}{Tabular}       & LR                              & 0.63±0.02 & 0.55±0.02      & 0.25±0.05 & 0.47±0.03 & 0.55±0.02 & 0.56±0.02                          & 0.70±0.02                     \\
                               & RF                              & 0.63±0.01 & 0.55±0.02      & 0.28±0.04 & 0.49±0.02 & 0.55±0.01 & 0.56±0.01                          & 0.71±0.01                     \\
                               & NN                              & 0.58±0.03 & 0.48±0.03      & 0.26±0.04 & 0.44±0.02 & 0.50±0.02 & 0.50±0.02                          & 0.63±0.02                     \\ \hline
\multirow{3}{*}{\vbox{\hbox{\strut Graph} \hbox{\strut ({\tt SPHN-tr})}}}     & TransE                          & 0.49±0.04 & 0.40±0.10      & 0.02±0.04 & 0.30±0.03 & 0.40±0.03 & 0.43±0.02                          & 0.50±0.01                     \\
                               & RDF2Vec                         & 0.50±0.05 & 0.39±0.14      & 0.01±0.02 & 0.30±0.03 & 0.39±0.04 & 0.44±0.02                          & 0.49±0.01                     \\
                               & RGCN3+lit                         & \textbf{0.84±0.01} & \textbf{0.76±0.02} & \textbf{0.64±0.08} & \textbf{0.75±0.03} & \textbf{0.75±0.02} & \textbf{0.78±0.01}                 & \textbf{0.91±0.01}                     \\ \hline
\end{tabular}%
}
\end{table}

\subsection{SPHN \textit{vs.} CARE-SM ontologies}
Similarly to the first experiment, TransE, Rdf2Vec and RGCN+lit are considered, but here applied to a KG instantiating either the SPHN or CARE-SM ontology. The model configuration and hyperparameter settings are the same as the first experiment. 
Because CARE-SM uses more predicates thus longer paths to connect patients with their features, we conducted an additional experiment on CARE-SM KG that consists in using RGCN with five layers. 
This aims to check if five-hop neighbors enable to capture enough information to predict the outcome. 
The obtained performance is reported in Table \ref{tab:sphncare-comparison}. TransE and RDF2Vec showed relatively weak performance on both KG. 
RGCN on SPHN showed the best performance (AUC=0.91). For CARE-SM, all models have difficulty on predicting the outcomes. Using this ontology, we note that RGCN with three layers is not performing better than TransE or RDF2Vec (AUC=0.50). 
Increasing the number of layers to 5 did not improve the performance (AUC=0.50).
We note that all the models barely predict the {\em Death} outcome. 

\begin{table}[ht]
\caption{The performance comparison of SPHN ({\tt SPHN-ts}) and CARE-SM ({\tt CARESM$^*$-ts}). RGCN3+lit and RGCN5+lit refers to RGCN+lit with 3 and 5 layers, respectively. {\tt CARESM$^*$} is the variant of CARE-SM. See Section \ref{subsec:graph-rep-of-clinic} for more details.}
\label{tab:sphncare-comparison}
\resizebox{\textwidth}{!}{%
\def\arraystretch{1.5}
\begin{tabular}{llccccccc}
\hline
\multirow{2}{*}{\textbf{KG}} & \multirow{2}{*}{\textbf{Model}} & \multicolumn{5}{c}{\textbf{F1-score}}                     & \multirow{2}{*}{\textbf{Accuracy}} & \multirow{2}{*}{\textbf{AUC}} \\ \cline{3-7}
                             &                                 & BackHome  & Rehab     & Death     & Macro     & Weighted  &                                    &                               \\ \hline
\multirow{3}{*}{\tt {SPHN-ts}}        & TransE                          & 0.51±0.07 & 0.33±0.16 & 0.02±0.04 & 0.29±0.04 & 0.37±0.05 & 0.43±0.02                          & 0.50±0.01                     \\
                             & RDF2Vec                         & 0.49±0.04 & 0.42±0.09 & 0.01±0.03 & 0.30±0.02 & 0.40±0.02 & 0.44±0.01                          & 0.50±0.02                     \\
                             & RGCN3+lit                          & \textbf{0.83±0.02} & \textbf{0.76±0.02} & \textbf{0.66±0.08} & \textbf{0.75±0.03} & \textbf{0.78±0.02} & \textbf{0.78±0.02}                          & \textbf{0.91±0.01}                     \\ \hline
\multirow{4}{*}{{\tt CARESM$^*$-ts}}     & TransE                          & 0.47±0.04 & 0.44±0.04 & 0.02±0.03 & 0.31±0.01 & 0.40±0.01 & 0.43±0.01                          & 0.49±0.01                     \\
                             & RDF2Vec                         & 0.51±0.07 & 0.38±0.11 & 0.00±0.00 & 0.29±0.02 & 0.39±0.03 & 0.44±0.02                          & 0.50±0.01                     \\
                             & RGCN3+lit                          & 0.53±0.08 & 0.30±0.17 & 0.00±0.00 & 0.28±0.04 & 0.37±0.05 & 0.44±0.01                          & 0.50±0.02                     \\
                             & RGCN5+lit                          & 0.48±0.08 & 0.30±0.19 & 0.00±0.00 & 0.26±0.04 & 0.34±0.05 & 0.44±0.05                          & 0.50±0.01                     \\ \hline
\end{tabular}%
}
\end{table}

\subsection{Various time modeling}

In this experiment, we compare predictive performance of graphs associated with various modeling of time as listed in Section \ref{subsec:graph-rep-of-clinic}. 
All experiments are conducted using the RGCN+lit model, except with no literals where it uses RGCN. Without literals, the model performs poorly, and under the standard approach from tabular data. The AUC increases about 33\% on average when the literals are added. This increase mainly come from a better prediction of the {\em Death} outcome. When the timestamps are added to {\tt SPHN-nt}, the AUC increases of 7\% on average. Adding the time relations also increases AUC of a 7\%. Adding both temporal information showed similar results as well. 
With saturation, we observe a slight incrase of performance for the {\em Death} outcome, though the overall performance is similar to {\tt SPHN-tsr}. Eventually, adding temporal information give the model better prediction, but the type of time modelling and the level of saturation did not make a significant difference.

\begin{table}[ht]
\caption{The Performance of RGCN+lit on SPHN with various temporal information and  modeling. RGCN without literals is applied to {\tt SPHN-nl}.}
\label{tab:time-info}
\resizebox{\textwidth}{!}{%
\def\arraystretch{1.5}
\begin{tabular}{lccccccc}
\hline
\multirow{2}{*}{\textbf{KG}} & \multicolumn{5}{c}{\textbf{F1-score}}                     & \multirow{2}{*}{\textbf{Accuracy}} & \multirow{2}{*}{\textbf{AUC}} \\ \cline{2-6}
                                & BackHome  & Rehab     & Death     & Macro     & Weighted  &                                    &                               \\ \hline
{\tt SPHN-nl}                         & 0.64±0.03 & 0.46±0.11 & 0.05±0.07 & 0.38±0.06 & 0.49±0.06 & 0.53±0.04                          & 0.64±0.06                     \\
{\tt SPHN-nt}                         & 0.75±0.02 & 0.65±0.02 & 0.55±0.06 & 0.65±0.02 & 0.68±0.01 & 0.68±0.01                          & 0.85±0.01                     \\
{\tt SPHN-ts}                         & 0.83±0.02 & \textbf{0.76±0.02} & 0.66±0.08 & 0.75±0.03 & 0.78±0.02 & 0.78±0.02                          & \textbf{0.91±0.01}                     \\
{\tt SPHN-tr}                         & \textbf{0.84±0.01} & \textbf{0.76±0.02} & 0.64±0.08 & 0.75±0.03 & 0.75±0.02 & \textbf{0.78±0.01}                 & \textbf{0.91±0.01}                     \\
{\tt SPHN-tsr}                        & 0.83±0.02 & \textbf{0.76±0.02} & 0.66±0.04 & 0.75±0.02 & \textbf{0.78±0.01} & \textbf{0.78±0.01}                          & \textbf{0.91±0.01}                     \\
{\tt SPHN-sat1}                       & 0.83±0.01 & \textbf{0.76±0.02} & 0.64±0.06 & 0.75±0.02 & \textbf{0.78±0.01} & \textbf{0.78±0.01}                          & \textbf{0.91±0.01}                     \\
{\tt SPHN-sat2}                       & 0.83±0.01 & \textbf{0.76±0.02} & \textbf{0.68±0.05} & \textbf{0.76±0.02} & 0.78±0.02 & 0.78±0.02        & \textbf{0.91±0.01}                     \\ \hline
\end{tabular}%
}
\end{table}

\section{Discussion}
\label{sec:discussion}

The comparative analysis reveals first that the RGCN+lit model outperforms baseline methods on tabular data, or other KGE approaches in terms of accuracy and F1-score. We hypothesis that the multi-relational modeling associated with the multiple RGCN-layers makes the model capable of aggregating information from multiple hop of neighbors to the patient node, what cannot be achieved from single relational modeling with tabular data, or with TransE or RDF2Vec which only partially consider the multi-hop neighbors.
Second, we observed that the choice of the schema for individual KG impacts the performance. In particular in our setting, the more compact and patient-oriented schema of SPHN is more favorable than the one of CARE-SM KGE approaches. This could be explained by the longer distance between patient and feature nodes in CARE-SM, however, increasing the size of the neighbor to five-hops did not solve the issue. However, we note that we did not expend further the number of hops for computational cost. 
A limitation to the conclusion relies on the fact that we considered only SPHN and CARE-SM as two prototypical ontologies. A more complete study would have considered concurrent models such as RDF-FHIR, PhenoPackets and others.
Third, adding time information helps RGCN+lit in classifying properly patient nodes, but we were enable to observe any difference in performance associated with various time modeling choices.  
Here we acknowledge that we focus on rather simple time modeling and that more complex scenarios exist such as those associated with dynamic graphs.





Overall, our study illustrates that KGE such as RGCN+lit represents a promising approach for predictive modeling in healthcare. However we note a strong class disbalance in our dataset (44.14\%, 43.33\% and 12.53\% for \textit{BackHome}, \textit{Rehabilitation}, \textit{Death}, respectively), reflecting the real world. This might explain to some extent the difficulty for most of the approaches to predict the
\textit{Death} class. It is further possibly mitigated by over-sampling techniques, such as Synthetic Minority Over-sampling \cite{chawla2002smote}. Also our study considers solely the task of node classification, whereas prediction could have been modeled as a link prediction or triple classification problem. More investigation would be necessary to assess if our conclusions stand in the context of other learning tasks. This could necessitate the consideration of problem classically associated with KGE such as negative sampling \cite{kamigaito2022comprehensive}.
Additionally, we observed that including literal embeddings in the model improved the performance. In this study, our model focused on the simple embedding of numerical literals, including timestamps. However, we plan to develop a model capable of handling multi-modal literals, such as a combination of text and numerical literals \cite{gesese2021survey}.

Another important position we took is between transductive and inductive learning approaches \cite{Xu2020Inductive}. We follow a transductive learning approach, which mean that node classification is made on the basis of the set of entities that has potentially been seen during training. Most of the KGE models are based on a transductive learning approach, because the model learns the complex relational information within a whole large graph. In contrast, inductive learning attempts to generalize the model to new entities and relations not presented in the training graph. For instance, it is representing the entities or relations based on the combination of observed entities or relations in the training graph \cite{galkin2022nodepiece}. Inductive approach can be adapted to the scenario where new patient information and medical events continuously emerges. This remains a future work of our study.

The effective evaluation of KG embedding models is crucial for advancing the field and ensuring that the models developed are robust, accurate, and useful in practical applications \cite{Gastinger2022}. But it suffers from a lack of standardized benchmarks and reference representations of the temporal dimension. In this study, we particularly aimed at advancing this agenda, by proposing a real-word task, a shared dataset and well documented baseline experiment, which will serve as a baseline for prediction modeling with graph data. 



An important area for future research is evaluating the impact of individual patient features on prediction outcomes. By analyzing the relative importance of different features, such as specific medical conditions or demographic factors, researchers can refine their models to focus on the most predictive attributes. This targeted approach can enhance the model's efficiency and ensure that the most relevant clinical information is prioritized in decision-making.
While at the moment, efforts are underway to obtain a small subset of a real dataset to test these models and generate predictions based on real patient data thanks to the collaboration with the 
Nantes
Hospital, in the longer term, the ultimate goal is to achieve clinical validation of the predictive models. This step involves applying the models to actual patient data and rigorously assessing their performance in a clinical setting. Clinical validation is critical for ensuring that the models are theoretically sound and practically useful in improving patient outcomes. This research phase will require close collaboration with healthcare providers and institutions to test the models in real-world environments and to gather feedback for further refinement.

In conclusion, while this study has made significant strides in demonstrating the potential of knowledge graph embeddings for predicting patient outcomes, a substantial amount of work remains to be done. By continuing to refine the models, enhance the realism of synthetic data, and pursue clinical validation, future research can build on these foundations to develop precise and clinically relevant predictive tools that ultimately enhance patient care and treatments.

\section*{Credits}
\subsubsection*{Acknowledgments} 
This work is supported by 
CombO (Health Data Hub) project and 
the Agence Nationale de la Recherche under the France 2030 program, 
ANR-22-PESN-0007 ShareFAIR, and ANR-22-PESN-0008 NEUROVASC.

\subsubsection*{Competing interests}
The authors have no competing interests to declare that are relevant to the content of this article.
%
%
%
\bibliographystyle{splncs04}
\bibliography{main}
\newpage
\section*{Appendix}
\appendix
\section{List of features of our synthetic dataset}
The list of clinical features in our synthetic data is shown in Table \ref{tab:list_features}. It includes 22 non-temporal features (4 numerical, 6 categorical, 12 binary) and 8 temporal features (named events in this manuscript). Note that events correspond to the time between hospital admission and the first occurrence of the event (\textit{e.g.}, the first administration of Nimodipine).

\begin{table}[ht]
\centering
\caption{List of clinical features.}
\label{tab:list_features}
\begin{tabular}{lll}
\hline
\textbf{Type}                                                                  & \textbf{Name}          & \textbf{Description}                    \\ \hline
\multirow{4}{*}{Numerical}                                                     & hospital\_stay\_length & Number of days in the hospital          \\
                                                                               & gcs                    & Glasgow coma score                                     \\
                                                                               & act\_nb               & Number of medical acts                  \\
                                                                               & age                    & Patient's age                                     \\ \hline
\multirow{6}{*}{Categorical}                                                   & gender                 & Gender                                  \\
                                                                               & entry                  & Mode of entry (\textit{e.g.}, emergency room)                           \\
                                                                               & entry\_code            & Code of entry                           \\
                                                                               & ica.y                    & Intracranial aneurysm (ICA) location and size \\
                                                                               & ica\_treatment                    & ICA treatment (\textit{e.g.}, endovascular or surgical)                           \\
                                                                               & ica\_therapy           & Calcium channel blockers therapy                            \\ \hline
\multirow{12}{*}{Binary}                                                       & fever                  & Fever                                   \\
                                                                               & o2\_clinic             & Decrease of oxygen, indirect measure                \\
                                                                               & o2                     & Deacrease of oxygen, blood measure                          \\
                                                                               & hta                    & Hypertension                            \\
                                                                               & hct                    & Hypercholesterolemia                              \\
                                                                               & smoking              & Smoking                                 \\
                                                                               & etOH                   & Alcohol consumption                                    \\
                                                                               & diabetes                & Diabetes                                \\
                                                                               & headache               & Headache                                \\
                                                                               & unstable\_ica               & Hemodynamic instability                 \\
                                                                               & vasospasm             & Vasospasm                               \\
                                                                               & ivh                    & Intraventricular hemorrhage             \\ \hline
\multirow{8}{*}{Event}                                                         & nimodipine             & Nimodipine                              \\
                                                                               & paracetamol            & Paracetamol use                            \\
                                                                               & nad                    & Norepinephrin use                                     \\
                                                                               & corotrop               & Milrinone use                               \\
                                                                               & morphine               & Morphine use                               \\
                                                                               & dve                    & External ventricular drainage                                     \\
                                                                               & atl                    & Percutaneous transluminal angioplasty                                     \\
                                                                               & iot                    & Orotracheal intubation                                     \\ \hline
\end{tabular}
\end{table}

\section{Statistics of Datasets}
The statistics of KGs, SPHN and CARE-SM.
\begin{table}[ht]
\centering
\caption{Statistics of Datasets}
\label{tab:dataset}
\resizebox{0.8\textwidth}{!}{%
\def\arraystretch{1.5}
\begin{tabular}{lcccc}
\hline
\textbf{Dataset} & \textbf{\# Entities} & \textbf{\# Relations} & \textbf{\# Literals} & \textbf{\# Triples} \\ \hline
SPHN             & 295,307              & 15                    & 36,415               & 1,127,467           \\
CARE-SM$^*$          & 576,733              & 13                    & 24,766               & 1,754,505           \\ \hline
\end{tabular}%
}
\end{table}

\vspace{7cm}

\end{document}